\documentclass[english, 10pt]{article}
\usepackage{authblk}
\usepackage{multicol}
\usepackage[a4paper]{geometry}
\usepackage[utf8]{inputenc}
\usepackage[T1]{fontenc}
\usepackage{babel}
\usepackage{fancyhdr}
\usepackage{graphicx}
\usepackage{amsmath,amsfonts,amssymb}
\usepackage{listings}
\usepackage{booktabs}
\usepackage[T1]{fontenc}
\usepackage{listings}
\usepackage{color}
\usepackage{hyperref}
\usepackage{lineno}
\usepackage{csquotes}
\hypersetup{colorlinks=true, linkcolor=blue, urlcolor=blue, citecolor=blue, pdfborder={0 0 255}}
\usepackage[natbib=true,style=numeric,sorting=none]{biblatex}
\DeclareUnicodeCharacter{0301}{\'{e}}
\usepackage{colortbl}
\usepackage{url}
\usepackage{caption}
\usepackage[toc, page]{appendix}
\usepackage{subcaption}
\usepackage{dirtytalk}
\usepackage{lineno}
\usepackage{setspace}
\usepackage{lipsum}
\usepackage{bm}

\newcommand*\samethanks[1][\value{footnote}]{\footnotemark[#1]}

\title{\textbf{Efficient Feature Representations for Cricket Data Analysis using 
Deep Learning based Multi-Modal Fusion Model}}
\date{}

\author[1]{Souridas Alaka\thanks{Equal contribution}}
\author[1]{Rishikesh Sreekumar\samethanks[1]}
\author[1]{Hrithwik Shalu\thanks{Corresponding author:- Hrithwik Shalu - \href{mailto:ae18b116@smail.iitm.ac.in}{\texttt {ae18b116@smail.iitm.ac.in}}}}

\affil[1]{Indian Institute of Technology Madras, India}
\makeatletter
\let\runtitle\@title
\runtitle

\begin{document}

\maketitle

\begin{abstract}
Data analysis has become a necessity in the modern era of cricket. Everything from effective team management to match win predictions use some form of analytics. Meaningful data representations are necessary for efficient analysis of data. In this study we investigate the use of adaptive (learnable) embeddings to represent inter-related features (such as players, teams, etc). The data used for this study is collected from a classical T20 tournament IPL (Indian Premier League). To naturally facilitate the learning of meaningful representations of features for accurate data analysis, we formulate a deep representation learning framework which jointly learns a custom set of embeddings (which represents our features of interest) through the minimization of a contrastive loss. We base our objective on a set of classes obtained as a result of hierarchical clustering on the overall run rate of an innings. It’s been assessed that the framework ensures greater generality in the obtained embeddings, on top of which a task based analysis of overall run rate prediction was done to show the reliability of the framework.
\end{abstract}

\section{Introduction}
Cricket is a bat and ball game played between two teams. This was found in early 16th century. There are different formats in cricket which are the tests which consists of two innings for each team and played in five days, ODI which gives each team 50 overs to bat and finishes in one day and T20 format which gives 20 overs each for team. T20 was introduced in 2005 and became much popular among various audience. The main reasons were the small duration of the match and the rules which are more favourable to batsmen compared to other formats. Indian Premier League (IPL) is such a cricket event where different teams based on different cities in India compete each other. IPL is based on the T20 format of cricket. IPL started in 2008 and have been a celebration among cricket followers around the globe. IPL has been earning lot of profit across the years.Players from different countries have been playing for these teams which increases the entertainment value as well as quality of cricket. In the latest IPL season(2020) BCCI earned a total profit of Rs 4000 crore. Along with increasing popularity of IPL, the analysis of players as well as teams across various conditions have became important among various sections.

Due to various factors like money, fan following, broadcasting, the entertainment value from matches is very important. This can be dependent on team selection, historical data of ground as well as the players, head to head to matches etc. This analysis can be used to analyse the effect of each player in a team combination with respect to opposition as well as the ground. Another similar use case of the analysis is in betting. We can find out fantasy points for each players from the model and provide good combinations to use in the betting platforms. As the entertainment value of the match highly depends on the batting performances, this analysis can help the pitch curators on making the decisions which will help them to create a bigger fan following as well in better marketing.

An inherent limitation to the current machine learning literature for sports data analysis is the lack of feature representations for players that contains necessary information on inter-player and inter-team relations. In this study we explore the use of vector embedding representations and the possible mediums to evolve them efficiently. Since the task has an inherent data scarcity (in comparison to the natural scenarios in which vector embeddings are used widely), we formulate a siamese-network\cite{schroff2015facenet} based representational learning framework that could effectively handle these limitations. 

\section{Related Work}\label{Related Work}
Data analytics is used in various aspects of cricket such as run prediction, player performance evaluation, team management, strategy formation etc.. There is a great demand for those algorithms that can perform the above tasks. IPL is a cricket event which is short in format and a lot of money involved. In India the followers  of  cricket  are  also  followers of  statistical  records. Thus the analysis of a league like IPL becomes more 
important. The following are some studies related to cricket which are reported in literature.

\cite{ref2}This article is concerned with the simulation of one-day cricket matches. Given that only a finite number of outcomes can occur on each ball that is bowled, a discrete generator on a finite set is developed where the outcome probabilities are estimated from historical data involving one-day international cricket matches. The simulator allows a team to easily investigate the results of making changes to the batting and bowling orders.

\cite{ref1}This paper presents a mathematical model that can be used for prediction  of the results of the matches prior to the match based on the knowledge of past matches, playing eleven and the toss result. In this work, three different models have
been constructed based on three approaches. Machine learning techniques have been utilized with advantage for this purpose. The outcome of a match is predicted  by taking a majority vote of these three models.

\cite{ref3}In this paper, prediction of the performance of players as how many runs will each batsman score and how many wickets will each bowler take for both the teams is performed. Both the problems are targeted as classification problems where number of runs and number of wickets are classified in different ranges. Random Forest turned out to be the most accurate classifier for both the data-sets with an accuracy of 90.74\% for predicting runs scored by a batsman and 92.25\% for predicting wickets taken by  a bowler. 

\cite{lamsal2020predicting}A multivariate regression based solution is proposed to calculate points for each player in the league and the overall weight of a team is
computed based on the past performance of the players who have appeared most for the team. Multi-layer Perceptron model gave the highest  accuracy of 71.66\% in predicting the outcome of the match.

\cite{ref5}Proposed a novel Recurrent Neural Network model which can predict the win probability of a match at regular intervals given the ball-by-ball statistics.

\pagebreak

\section{The Proposed Method}\label{Method}

We split our overall methodology into three sub-components, initially we learn the Embedding representations of players independently of other features. Next we use the learned embeddings from the model along with other commonly available pre-match data features to predict the overall run-rate. Finally we include a separate branch in the network to analyse the corresponding pre-match pitch report to investigate it's impact on overall prediction metrics.

\subsection{Player Embedding Model}\label{player_emebed_Model}

The model composes of two separate embedding maps that represents the batting and bowling characteristics of each player. The embedding vectors are a normalized set of size 64. We mean-pool the embeddings from each map and pass the same through a fully connected layer with ReLU activation so as to obtain a latent representation of each team's characteristics. A joint representation of the batting and bowling team corresponding to an innings is obtained by concatenating the latent vectors so obtained from each branch. The joint representation so obtained is passed through a series of fully connected layers (with ReLU non-linearity) attached to a prediction or representational head. The player embedding model is trained independently so as to avoid induced bias from other data features. 

\begin{figure}[!htb]
    \centering
    \includegraphics[width=\textwidth]{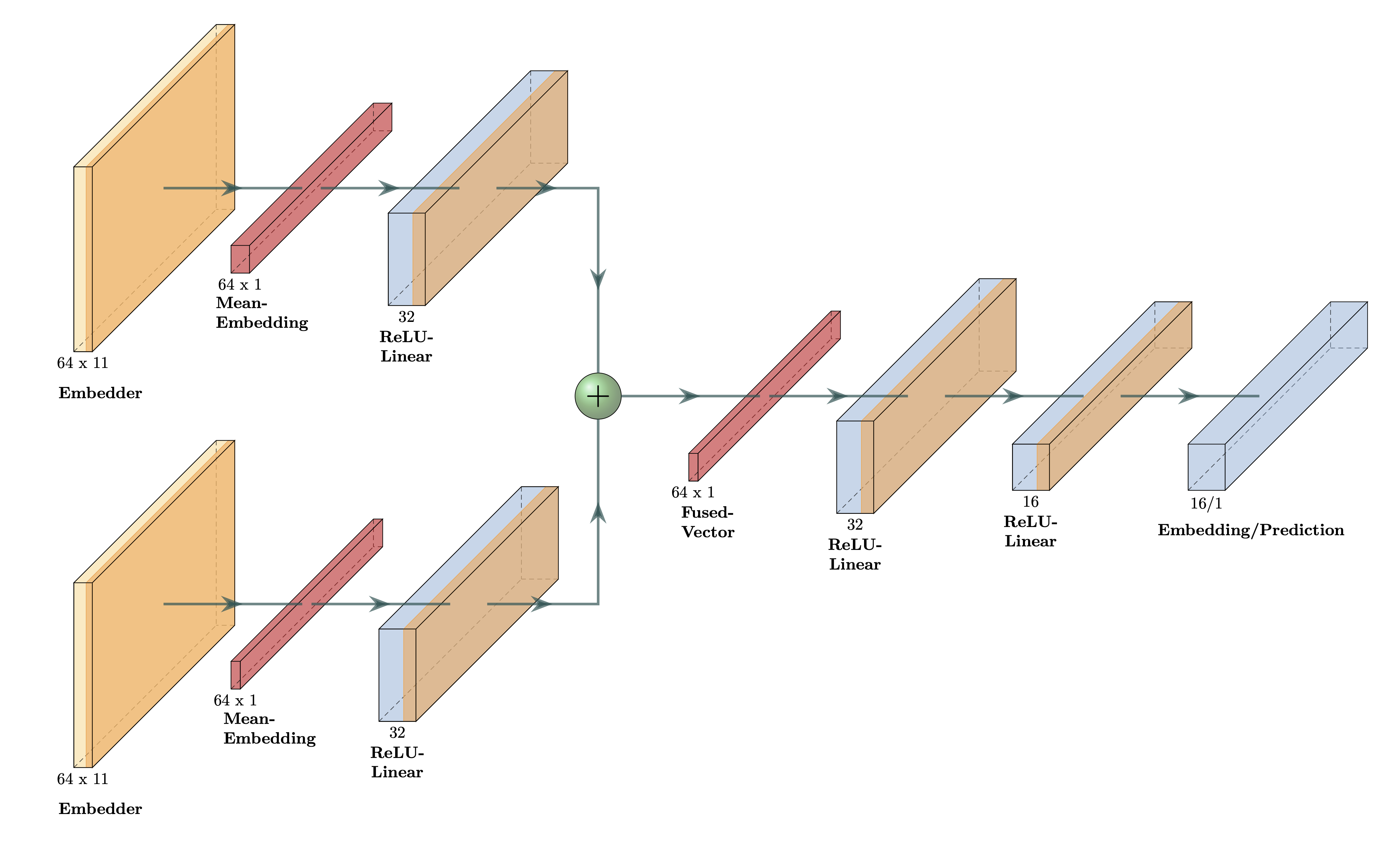}
    \caption{\textbf{Complete architecture of the player embedding model.} The final layer of the given model could change with the training setting used.}
    \label{fig:player_embed}
\end{figure}

\subsection{Prediction Model}\label{Pred_model}

Since player lineups aren't the only deciding factor in predicting match metrics, we investigate the impact of other commonly available match and timeline features on match metrics. In this section we build a prediction model that bases it's prediction on the analysis of joint representations formed from the latent vectors formed from commonly available pre match data and player representations. The bowling and batting embedding matrices used are obtained from the player embedding model \ref{player_emebed_Model}, the embedding representations are frozen during training. We increase the no of fully connected layers in each branch that processes the player embeddings as compared to the player embedding model, the same is done to evaluate the generality of the embedding representations formed.

\begin{figure}[!htb]
    \centering
    \includegraphics[width=1\textwidth, height=13cm]{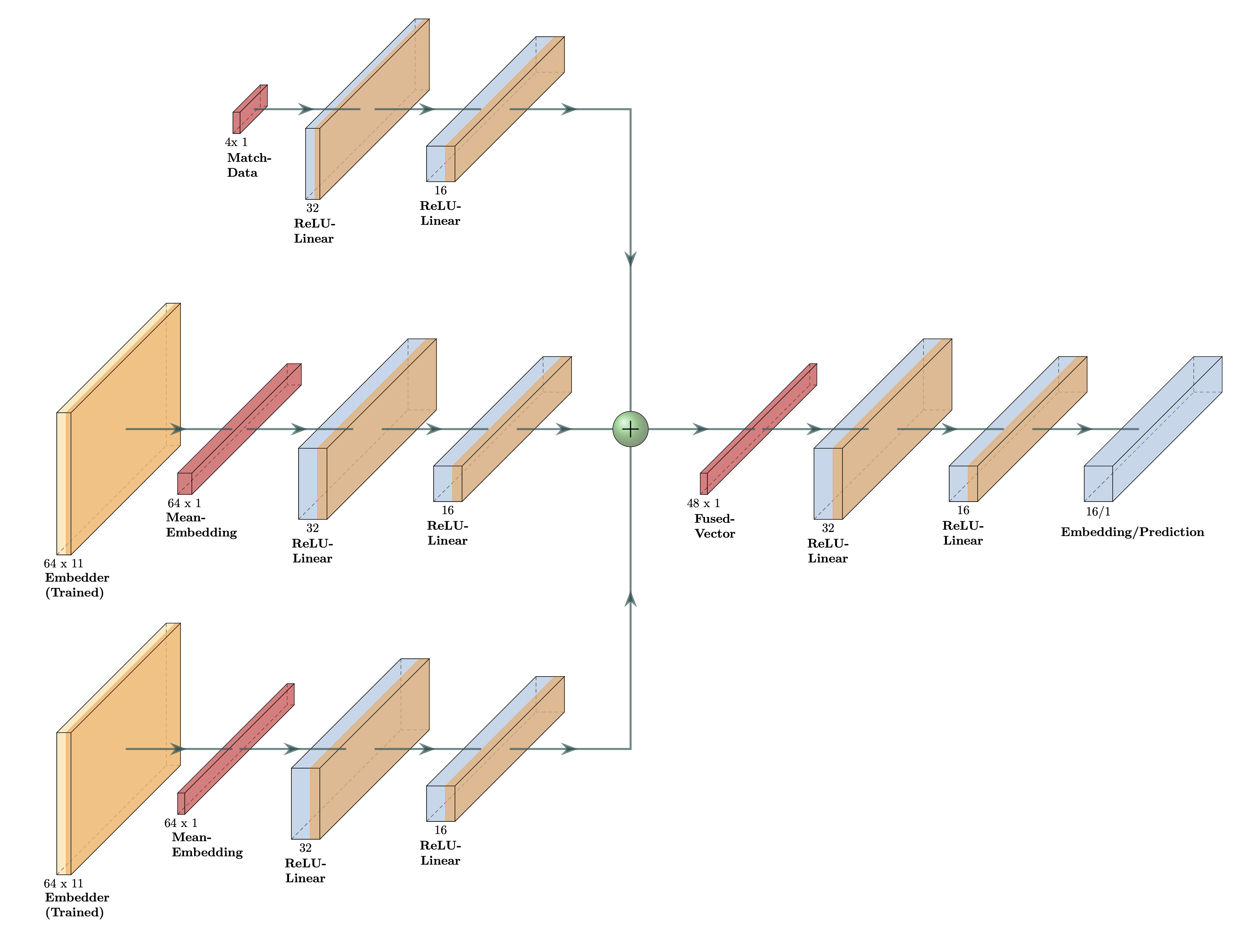}
    \caption{\textbf{Complete Architecture of the Prediction model}. The model is used as a medium for evaluating the effectiveness of the overall formulation of using embedding representations}
    \label{fig:prediction_model_1}
\end{figure}

\subsection{Prediction Model with pitch reports}\label{Pred_model_pitch}

Pitch reports are an essential source of pre-match information, that could prove vital in effective overall modelling . To evaluate the impact of the same we build a separate model which incorporates information from pitch reports into the joint representations formed. The pitch reports are prepossessed using a sentence vectorizer prior to model input.

\begin{figure}[!htb]
    \centering
    \includegraphics[width=1\textwidth, height=15cm]{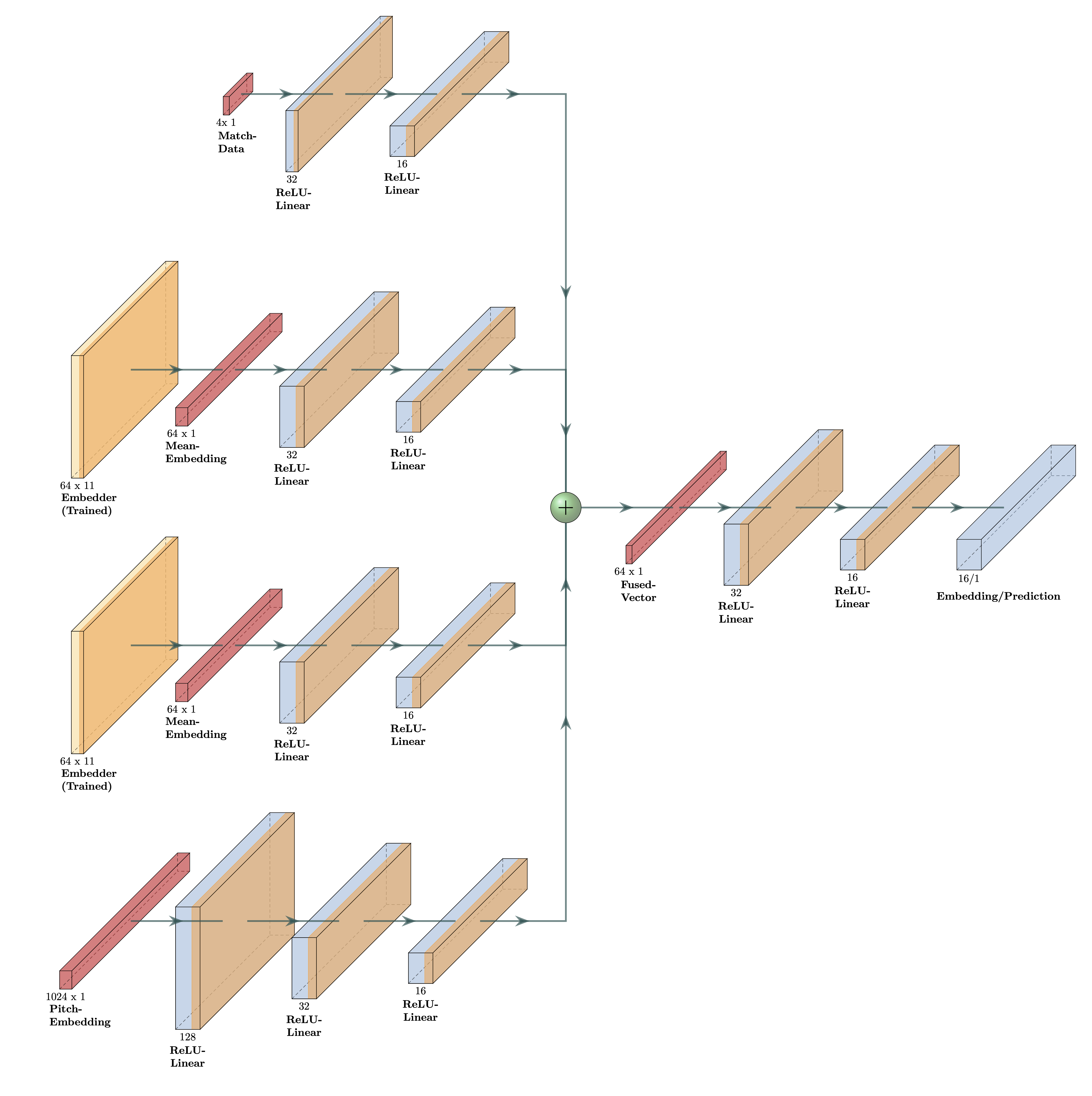}
    \caption{\textbf{Complete Architecture of the Prediction model with pitch reports}. A Joint representation is formed from the concatenation of latent vectors obtained from each of the different data branches.}
    \label{fig:prediction_model_2}
\end{figure}

\subsection{Representational Learning}

Apart from using the traditional cross-entropy loss to build the classifier, we experiment with a representational learning framework which generates meaningful data representations for classification by latent vector comparisons. The objective function used is the contrastive loss \cite{schroff2015facenet}, the standard form of which is as shown in Equation \ref{eq1}. 

\begin{equation}
L(W, (Y, \vec{X_1}, \vec{X_2})^i) = (1 - Y)L_S(D_W^i) + Y L_D(D_W^i)
\label{eq1}
\end{equation}

\noindent
$Y$ is the binary label that indicates if two data samples are acquired from a common category or not and $D_W$ is the parameterized distance function as shown in Equation \ref{eq2}.
\begin{equation}
D_W(\vec{X_1}, \vec{X_2}) = \left\|G_W(\vec{X_1}) - G_W(\vec{X_2}) \right\|_2
\label{eq2}
\end{equation}

\noindent
For our purposes we choose $D_W$ as the euclidean norm, $L_S$ = $D_W^i$, and $L_D$ as the $min(m - D_W^i, m)$. Where $m$ is the margin parameter. The form of the final loss function is  shown in Equation \ref{eq3}.

\begin{equation}
L(W, (Y, \vec{X_1}, \vec{X_2})^i) = (1 - Y)(D_W^i) + Y (min(m - D_W^i, m))
\label{eq3}
\end{equation}

\section{Experiments and Results}\label{Results}
\subsection{Dataset} \label{Dataset}
The data which contains the team lineup, venue, date and run-rate was collected from the official IPL webpage~\cite{iplt20}. It includes the innings wise data from the year 2012 to 2019. sportskeeda.com~\cite{pitch1}, espncricinfo.com~\cite{pitch2} and cricbuzz.com~\cite{pitch3} were the sources for the pitch report data. K-Means clustering~\cite{cluster} was performed on the run rate data to to divide into three classes. Elbow method~\cite{elbow} was used to find the optimal number of classes. Further Hierarchical clustering was done to obtain better class distribution, splitting the majority class into two sub classes. The following figure \ref{fig:picture} shows the  average dispersion across different number of clusters.

\begin{figure}[!htb]
    \centering
    \includegraphics[width=0.8\textwidth, height=6cm]{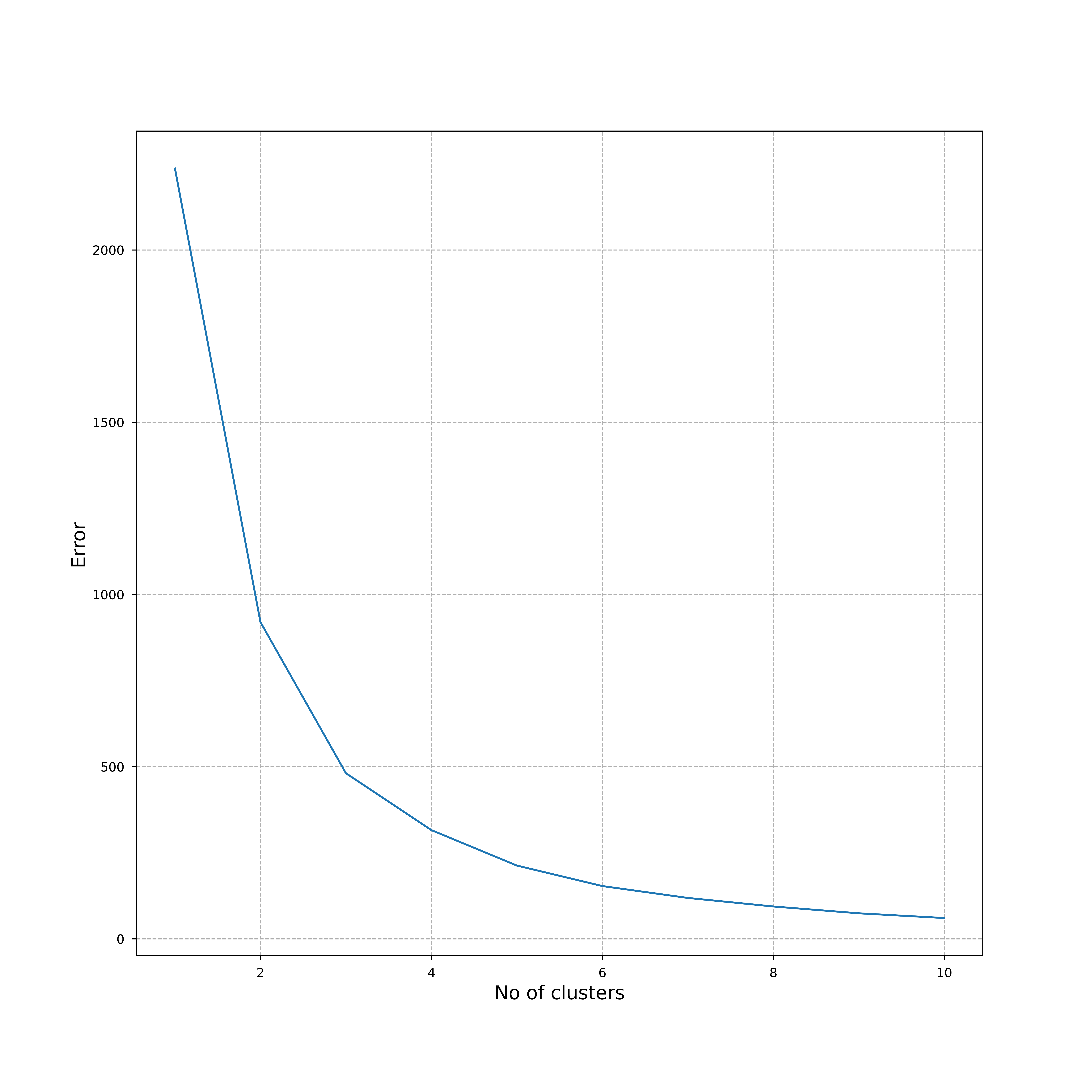}
    \caption{\textbf{The Elbow curve obtained for initial clustering}. The optimal initial no of clusters to be used can be infer-ed from the shown elbow curve and can be seen to be = '3'.}
    \label{fig:picture2}
\end{figure}

\noindent
Four classes were obtained after the hierarchical clustering of overall run rate per innings. The following pie diagram \ref{fig:picture} shows the individual class distributions and the centroid of respective clusters. The dataset so obtained has overall class balance.

\begin{figure}[!htb]
    \centering
    \includegraphics[width=0.7\textwidth, height=5cm]{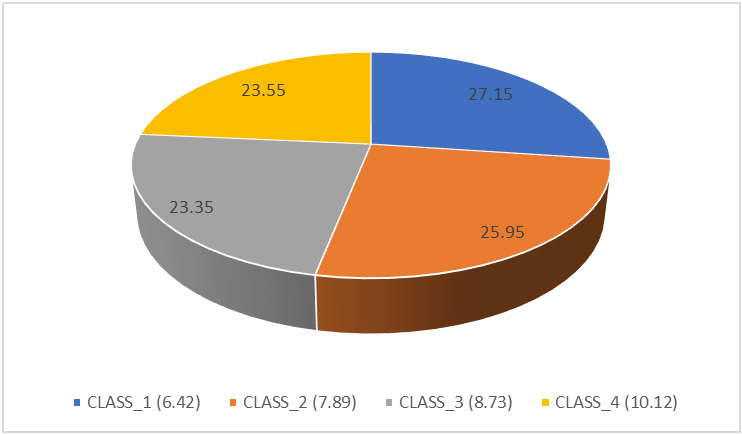}
    \caption{\textbf{Final class distribution}. The centroid of each corresponding class is as marked in the figure}
    \label{fig:picture}
\end{figure}

\noindent
The testing set is randomly sampled each time during k-fold cross validation and confidence intervals estimations. The overall data distribution used in this study remains invariant.

\subsection{Experimental Setup}

A separate testing set was created by randomly sampling 10 data points from each run-rate class obtained post hierarchical clustering. We conduct our experiments as two separate settings, Firstly we train the player embedding model by basing it's objective to predict the relevant class using the standard cross entropy loss. The trained set of embeddings so obtained are used as inputs during the training of the prediction model, the objective function used for the same is also cross-entropy. 

Next we train the embedding model using the contrastive loss to enforce meaningful representations, the embedding matrix so obtained is used to train a the prediction model whose objective is also the minimization of a contrastive loss. For evaluation we perform similarity analysis between the obtained representation of a test data point and the representation matrix corresponding to the train set.

Finally for both the settings given above, we add a separate branch of input in the prediction model so as to include the pre-match pitch report data. The pitch report is processed into a representational embedding using a pretrained sentence BERT model \cite{reimers2019sentence}. We thoroughly evaluate the model performance in both settings and the impact pitch-data has in predicting the overall run-rate class.

All experiments were conducted in the Pytorch framework with a CUDA backend on NVIDIA Tesla V100 GPU. The prediction model is trained in batch-sizes of 64 using the Adam optimizer. The learning rate used is 10\textsuperscript{-3}, the player and pitch embedding weights are frozen during the training of the prediction models.

\pagebreak

\subsection{Results}

\begin{figure}[!htb]
\begin{minipage}[b]{0.48\textwidth}
\includegraphics[width=\linewidth]{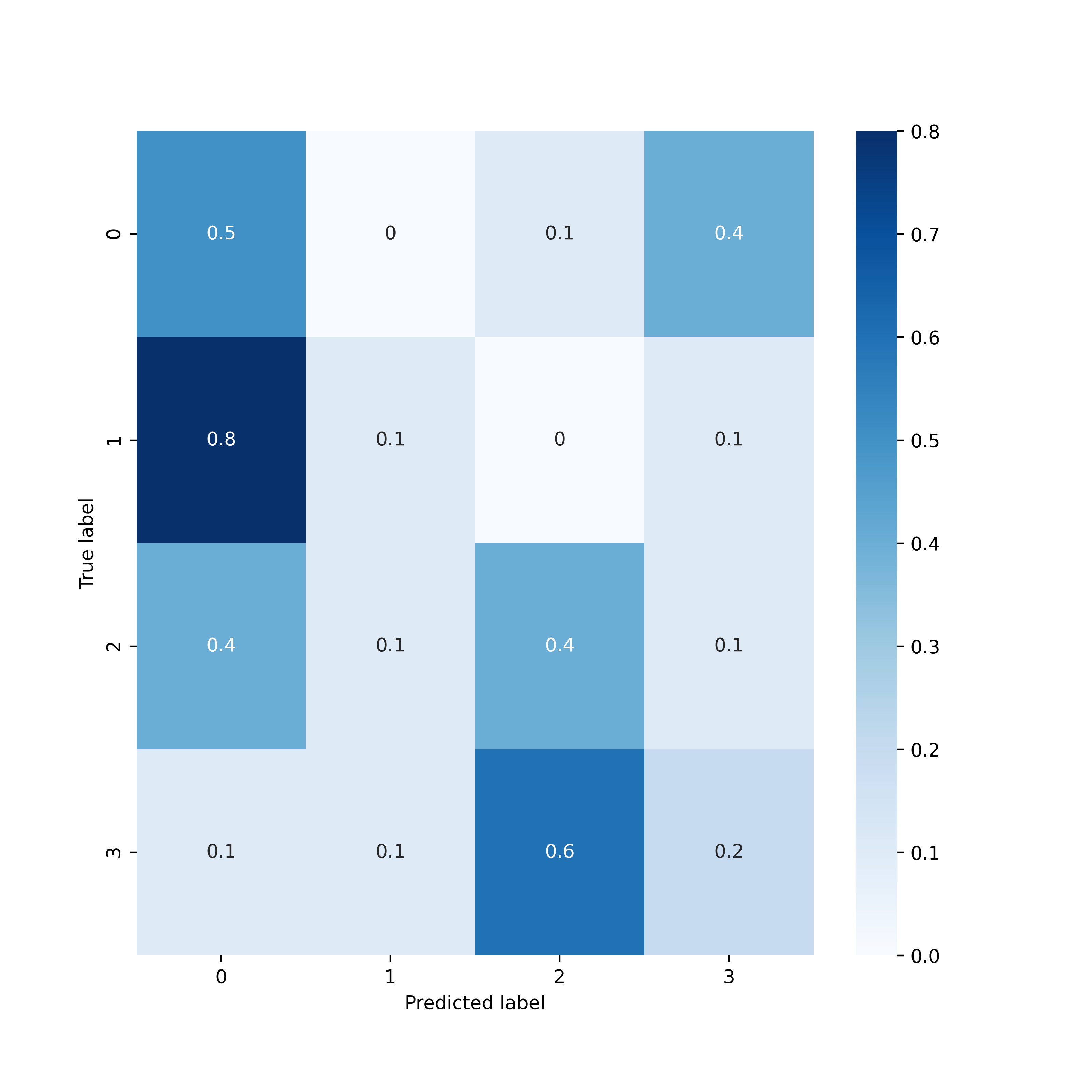}\\
\subcaption{Confusion matrix with pitch data}\label{fig:a0}
\end{minipage}%
\begin{minipage}[b]{0.48\textwidth}
\includegraphics[width=\linewidth]{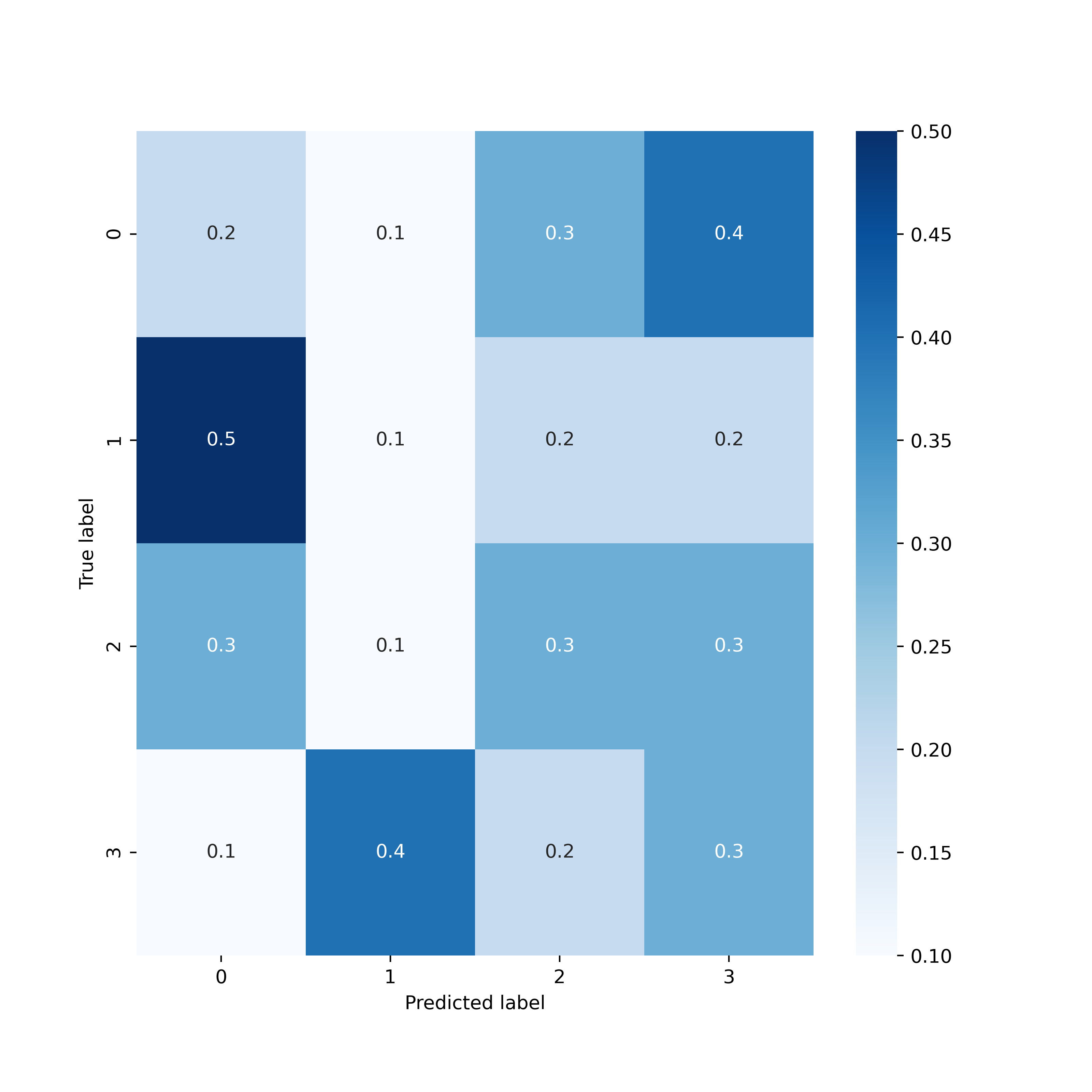}\\
\subcaption{Confusion matrix without pitch data}\label{fig:b0}
\end{minipage}

\caption{The shown confusion matrices shown are obtained after training the prediction models till point of divergence using the categorical cross entropy loss.}\label{fig5}
\end{figure}

\begin{figure}[!htb]
\begin{minipage}[b]{0.48\textwidth}
\includegraphics[width=\linewidth]{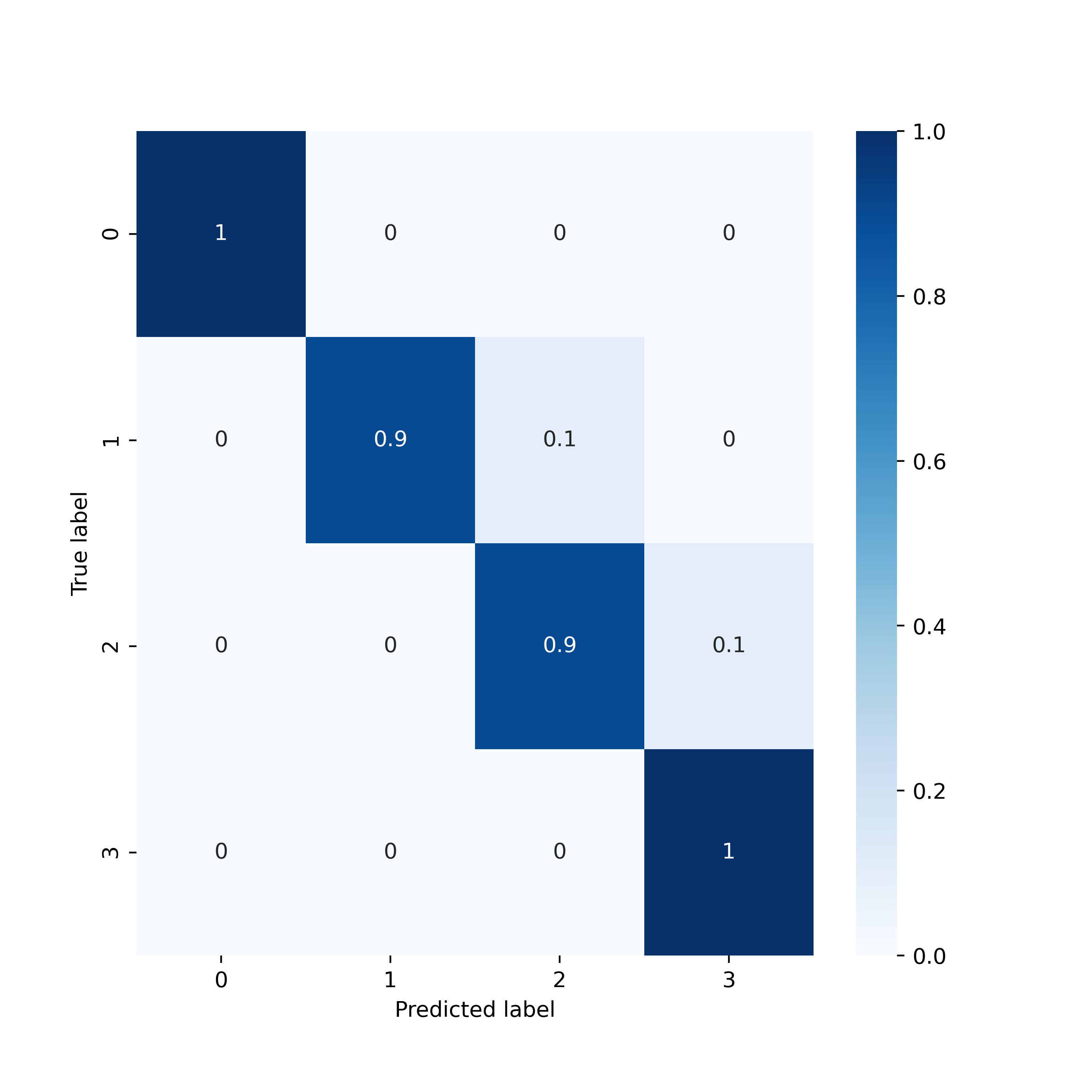}\\
\subcaption{Confusion matrix with pitch data}\label{fig:a1}
\end{minipage}%
\begin{minipage}[b]{0.48\textwidth}
\includegraphics[width=\linewidth]{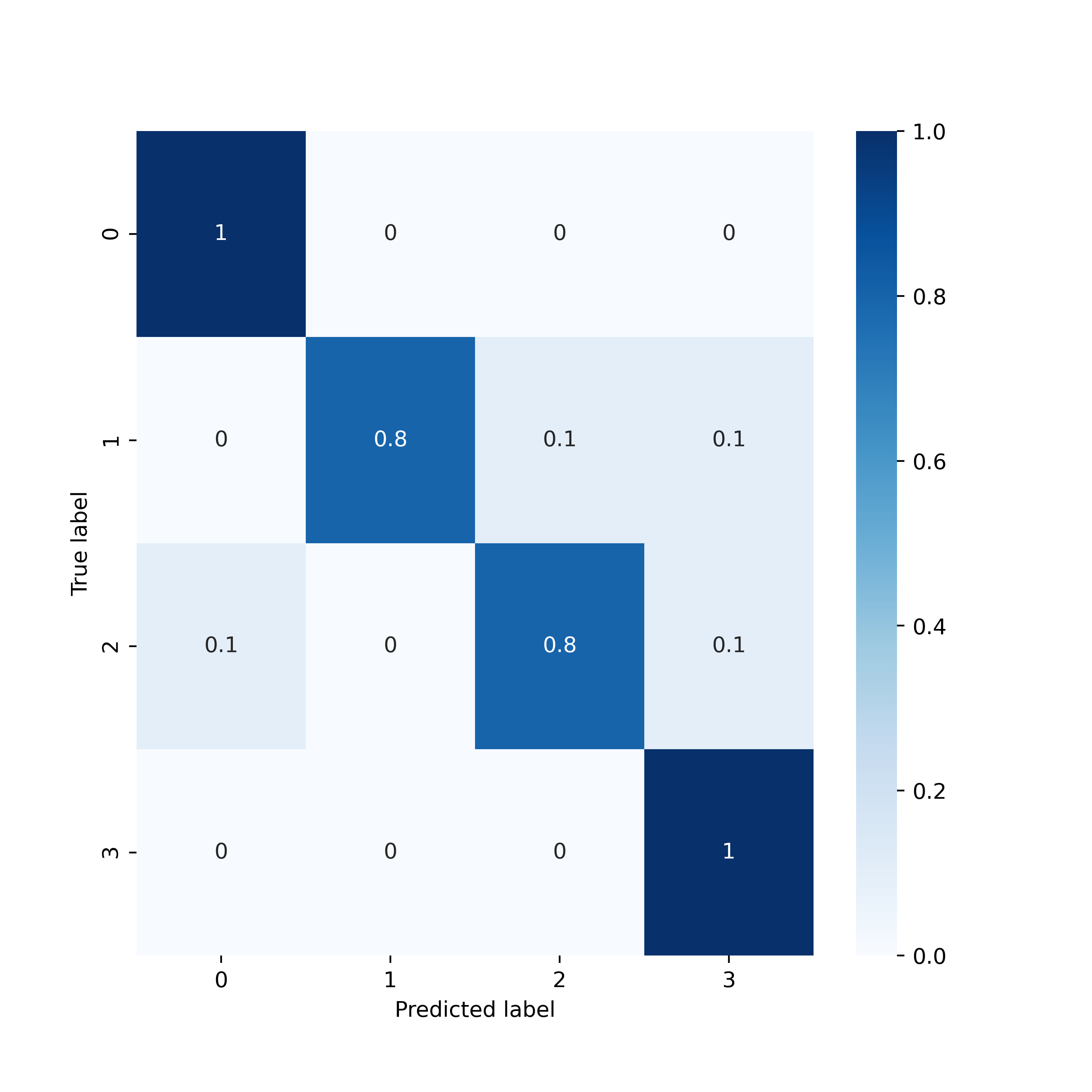}\\
\subcaption{Confusion matrix without pitch data}\label{fig:b1}
\end{minipage}

\caption{The shown confusion matrices shown are obtained after training the prediction models in the representational learning framework for 2000 epochs}\label{fig6}
\end{figure}

\noindent
The training setting that used cross-entropy loss acquired an accuracy of 22.5\% (CI -95\%) when pitch report data was not included (corresponding  confusion matrix shown in Figure \ref{fig:b0}), the model that based it's prediction with all of the prior's input and pitch report data achieved an accuracy of 30\% (CI -95\%) (corresponding  confusion matrix shown in Figure \ref{fig:a0}).
The representational learning framework showed a considerable increase in performance and achieved an accuracy of 90\% (CI -95\%) when pitch report data was not included (corresponding confusion matrix shown in Figure \ref{fig:b1}) and an accuracy of 95\% (CI -95\%) in the model setting that included pitch reports (corresponding confusion matrix shown in Figure \ref{fig:a1}).

\section{Discussion}\label{Discussion}
The setting that used cross-entropy loss achieved feeble accuracies as compared to the representation learning setting. We speculate the reason for the same being the inability of cross-entropy setting to learn meaningful vector embeddings for player feature representations. The model setting that included the pitch reports showed considerable increase in performance, the same maybe attributed by the similarities captured in the pitch embedding vectors. The cross-assessment of pitch properties could thus be speculated to be crucial for overall score-performance analysis. The prediction inaccuracies in intermediate classes could be associated with the lack of efficient input features to fully represent the match characteristics.

\section{Future Works}
Although the present work was focused entirely on efficient modeling of player representations we speculate that their might be overall bias to the current setting due to data scarcity. The semantic meanings of the player representations could be analysed and made robust by employing the same for better prediction tasks. We believe that the model architecture could be tuned further, maybe by introducing a convolutional mode of analysis for player embeddings and a relative weighted concatenation for joint representations.

\section{Conclusion}\label{Conclusion}
In this paper, we proposed a representational learning framework for optimal cricket data analysis. We observe that meaningful data representations could be obtained for players using the same. The performance of prediction models trained using cross entropy objective was proven to be feeble as compared to the proposed framework. We hope our work would act as a motivation for future research using  deep-representational framework and learnable input representations for sports analytics tasks.

\printbibliography

\section*{Code Availability}
The custom Python based code used in this study is available from the corresponding author upon reasonable request. Any commercial use including the distribution, sale, lease, license, or other transfer of the code to a third party, is prohibited. 

\end{document}